\documentclass[times, review, 10pt]{elsarticle}

\usepackage{array} 
 
\usepackage{booktabs} 
\usepackage{multirow}

\usepackage{amssymb}
\usepackage{amsmath}
\usepackage[numbers]{natbib}

\newcommand{\ie}{\textit{i.e.}}
\newcommand{\etal}{\textit{et. al.}}

\newcommand{\name}{SARTM}

\journal{Nuclear Physics B}
\usepackage[numbers]{natbib}
\def\tsc#1{\csdef{#1}{\textsc{\lowercase{#1}}\xspace}}
\tsc{WGM}
\tsc{QE}

\begin{document}

\begin{frontmatter}

\title {SARTM: Segment Any RGB Thermal Model with Language aided Distillation
}  





\author[1,2]{Dong Xing}
\author[1,2]{Jinhe Zhang}
\author[1,2]{Hang Yang}
\author[1,2]{Yuqing Wang}

\affiliation[1]{organization={Changchun Institute of Optics, Fine Mechanics and Physics, Chinese Academy of Sciences},
            city={Changchun},
            postcode={130033}, 
            state={Jilin},
            country={China}}
\affiliation[2]{organization={University of Chinese Academy of Sciences},
            city={Beijing},
            postcode={100049}, 
            country={China}} 
\cortext[1]{Corresponding author: Yuqing Wang}

\begin{abstract}
The recent Segment Anything Model (SAM) demonstrates strong instance segmentation performance across various downstream tasks. However, SAM is trained solely on RGB data, limiting its direct applicability to RGB-thermal (RGB-T) semantic segmentation. Given that RGB-T provides a robust solution for scene understanding in adverse weather and lighting conditions, such as low light and overexposure, we propose a novel framework, \name, which customizes the powerful SAM for RGB-T semantic segmentation. Our key idea is to unleash the potential of SAM while introduce semantic understanding modules for RGB-T data pairs.
Specifically, our framework first involves fine-tuning the original SAM by adding extra LoRA layers, aiming at preserving SAM's strong generalization and segmentation capabilities for downstream tasks. 
Secondly, we introduce language information as guidance for training our \name. 
To address cross-modal inconsistencies, we introduce a Cross-Modal Knowledge Distillation(CMKD) module that effectively achieves modality adaptation while maintaining its generalization capabilities. This semantic module enables the minimization of modality gaps and alleviates semantic ambiguity, facilitating the combination of any modality under any visual conditions. Furthermore, we enhance the segmentation performance by adjusting the segmentation head of SAM and incorporating an auxiliary semantic segmentation head, which integrates multi-scale features for effective fusion. 
Extensive experiments are conducted across three multi-modal RGBT semantic segmentation benchmarks: MFNET, PST900, and FMB. Both quantitative and qualitative results consistently demonstrate that the proposed SARTM significantly outperforms state-of-the-art approaches across a variety of conditions. Code and pre-trained weights can be found at https://github.com/wahaha-debug/SARTM.
\end{abstract}

\begin{keyword}
RGB-Thermal Semantic Segmentation, Segment Anything, Language Distillation
\end{keyword}
\end{frontmatter}

\section{Introduction}
In the field of autonomous driving, robust and reliable semantic scene understanding is critical for ensuring the safety of vehicle operations~\cite{Dehazed1}. RGB-thermal (RGBT) semantic segmentation technology provides a significant avenue for addressing the scene understanding challenges in adverse weathers and lighting conditions. For instance, during foggy or low-light scenarios, RGB cameras struggle with object recognition due to low visibility, while thermal infrared cameras can effectively detect objects by utilizing their thermal signatures~\cite{lv2024context}. By fusing information from both modalities, more accurate and robust semantic segmentation can be achieved in complex environments~\cite{MFFENet}. As a result, RGBT semantic segmentation has garnered considerable attention in recent years, with notable advancements in the field~\cite{sun2020fuseseg}.

\begin{figure}
  \begin{center}
     \includegraphics[width=\linewidth]{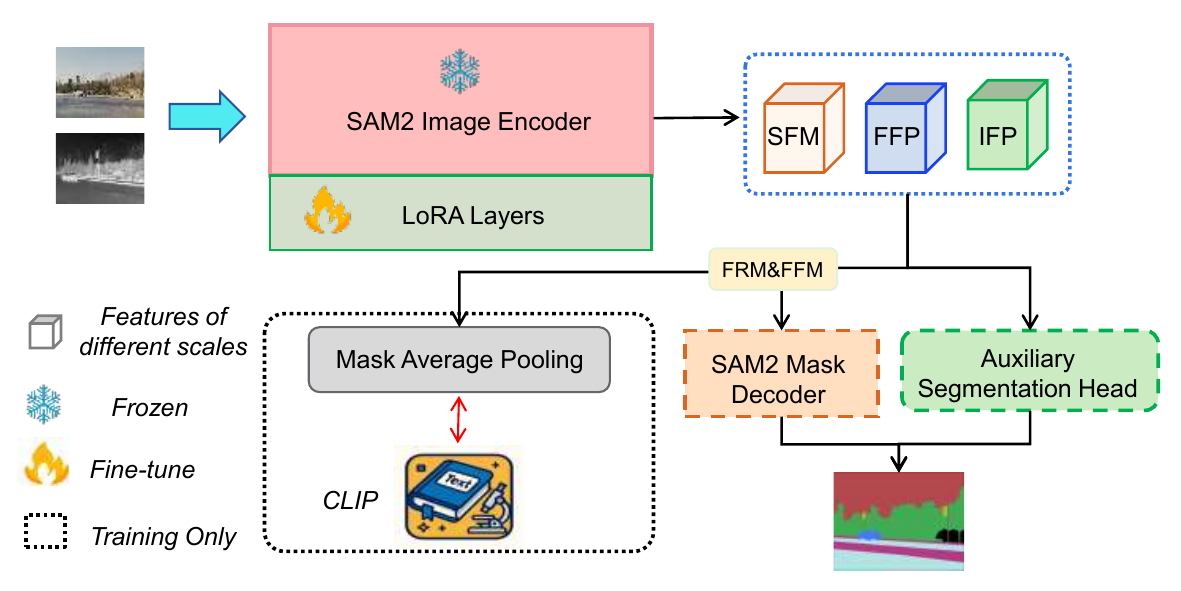}
     \vspace{-20pt}
     \caption{
     Overall framework of our proposed \name, consists of original SAM2 components and proposed language distillation part,including the Semantic Feature Map (SFM), Fine-Grained Feature Pyramid (FFP), and Intermediate-Resolution Feature Pyramid (IFP).}\label{fig:overall}
  \end{center}
\end{figure}

Recently, the Segment Anything Model (SAM)~\cite{kirillov2023segment} represents a breakthrough, particularly in RGB image segmentation. 
Despite SAM's impressive performance in single-modality segmentation tasks, its application in multi-modal segmentation remains a \textit{\textbf{unique challenge}}. Specifically, in multi-modal scenarios such as thermal infrared, the data from different modalities exhibit significant differences, and how to effectively integrate complementary information is still an open problem. The recently proposed SAM2 model~\cite{ravi2024sam} extends SAM’s capabilities by incorporating the temporal dimension to address challenges in video segmentation, such as motion, deformation, occlusion, and lighting changes. These advancements expand SAM's applicability to dynamic and multi-modal environments, but further research is needed to explore \textbf{\textit{how to effectively fuse cross-modal information while retaining SAM2's strong generalization ability}}.
Additionally, SAM not directly perform pixel-level fine-grained predictions, as it focuses more on segmenting object regions rather than precise pixel-level labeling. 


Although image fusion technology has wide applications, existing methods heavily rely on visual features, such as texture, contrast, and pixel registration, while neglecting deeper semantic information~\cite{MsgFusion}. As a result, deeper semantic information inherent within the images is often neglected. A pressing challenge is to effectively leverage the deeper, non-visual semantic features present in the images. 
Inspired by the successes of multi-modal vision-language model~\cite{han2023imagebind}, we explore a novel approach under the guidance of MVLM,\ie, CLIP~\cite{radfordlearning}, addressing the challenge by learning modality-agnostic representations and opening a new direction in RGBT fusion.

    To address the aforementioned challenges including: \textcircled{1} adapt SAM2 to multi-modal domain, \textcircled{2} introduce language as guidance for RGBT, and \textcircled{3} equip SAM2 semantic understanding ability, 
    we present a novel framework SARTM based on the SAM2 architecture. 
    As illustrated in Figure\ref{fig:overall}, The SAM2 architecture requires modality-specific fine-tuning to improve segmentation performance across various modalities.
    Built upon SAM2 model, we further propose: \textbf{\textcircled{1}}. to first adapt SAM2 which is solely trained with RGB data, we apply the Low rank adaptation (LoRA) layers to the image encoder to facilitates effective modality-adaptation fine-tuning while preserving the generalization ability of SAM2's pre-trained knowledge; The image encoder processes the input visual modality and generates a semantic feature map (SFM). This map is then passed through the mask decoder's convolutional module, which creates two additional feature pyramids: the fine-grained feature pyramid (FFP) and the intermediate-resolution feature pyramid (IFP).These pyramids, along with the SFM, enhance the model’s \textit{\textbf{spatial and semantic representation capacity}}. \textbf{\textcircled{2}}. aiming at utilizing the pretrained knowledgeable large vision language models, we integrate language information as guidance into the language-aided distillation module. This module allows the model to \textbf{\textit{better understand the intrinsic semantic correlation within the RGBT data}}, and finally improve the semantic segmentation accuracy in even complex and challenging environments. \textbf{\textcircled{3}}. to introduce the SAM2 which is only capable of instance segmentation to semantic segmentation, we modify the original mask decoder of SAM2 and further incorporate a auxiliary semantic decoder. Aiming at \textbf{\textit{effectively integrating multi-scale features}} is essential for improving segmentation accuracy in dynamic and diverse environments, we upgrade the SAM2 segmentation pipeline by incorporating multi-scale feature extraction and fusion mechanisms. Specifically, we augment the original segmentation head with an auxiliary head to leverage complementary information across multiple scales, thereby improving segmentation accuracy.

Extensive experiments are conducted on three benchmark datasets, including MFnet~\cite{MFNet}, PST900~\cite{PST900}, and FMB ~\cite{FMB}, demonstrate the outstanding performance of our framework in the RGB-T semantic segmentation task. As shown in Figure 1, our method achieves significant improvements on the PST900 dataset, while also demonstrating notable enhancements on the MFnet dataset.
Our contributions are as follows:

\noindent \textbf{(i)} We modify the SAM2 framework by integrating the LoRA layers into the multi-modal semantic segmentation task, investigating their potential for RGB-T image segmentation and adapting the framework to the multimodal domain.

\noindent \textbf{(ii)} We incorporate language descriptions into the training process by embedding explicit (language model-derived) textual guidance into the image fusion algorithm. This allows the model to better grasp the semantic context, enhancing segmentation accuracy in complex scenarios to support semantic segmentation.

\noindent \textbf{(iii)} We redesign the SAM2 segmentation pipeline by merging modified segmentation heads for multi-modal inputs and introducing an auxiliary segmentation head. This configuration enables efficient multi-scale feature fusion, leading to a significant improvement in segmentation precision.

\noindent \textbf{(iv)} Our method achieves state-of-the-art performance on three widely used multi-modal benchmarks, spanning synthetic to real-world scenarios, outperforming existing methods in segmentation accuracy and generalization across various dimensions.

\section{Related Works}
\subsection{RGB-T Semantic Segmentation.}
Multi-modal semantic segmentation often enhances performance by integrating the RGB modality with additional visual modalities containing complementary scene information, such as thermal and depth modalities. These supplementary modalities provide critical information for vision systems in diverse scenarios.
In RGB-thermal semantic segmentation, Ha \etal~\cite{MFNet}  introduced MFNet as a pioneering method for RGB-T semantic segmentation, employing a dual-stream architecture for feature extraction and cascading the fusion of these features. Sivakumar \etal~\cite{PST900} proposed a two-stage framework where the output of the first stage was fused with thermal and color images. However, these approaches primarily rely on simple fusion strategies, such as element-wise summation or concatenation, to capture cross-modal features, which can lead to redundancy by overlooking differences between cross-modal information. To address this, several studies have focused on designing specialized feature fusion operations. For instance, Lv \etal~\cite{lv2024context} proposed the Context-Aware Interaction Network (CAINet), which establishes complementary relationships between multi-modal features and long-term contexts in spatial and channel dimensions. Zhang \etal~\cite{zhangMRFS} introduced MRFS, where different modalities provide learning priors to each other. By contrast, \textit{we improve RGB-thermal semantic segmentation by leveraging SAM for robust segmentation and LoRA for fine-tuning, enhancing cross-modal feature alignment and reducing redundancy}.

\subsection{SAM for Semantic Segmentation.} 
The Segment Anything Model (SAM)~\cite{kirillov2023segment} is a versatile segmentation framework that achieves unprecedented performance in image segmentation by leveraging a combination of large-scale, diverse datasets and robust vision models. SAM has catalyzed advancements across various subfields of computer vision. Its architecture consists of three core components: an image encoder, a prompt encoder, and a mask decoder.
SAM has demonstrated its effectiveness in multiple domains, including medical imaging~\cite{li2026domain} and 3D reconstruction~\cite{NTO3D}. In our work, we pioneer the application of SAM to this domain by training it on RGB-X semantic segmentation tasks, marking the first exploration of SAM in this context. We adapt SAM to RGB-T semantic segmentation by fine-tuning it with LoRA, enhancing cross-modal feature alignment and improving segmentation accuracy.

\subsection{Vision-Language Model.}
Recently, vision-language multi-modal learning~\cite{LearningRadford}, has emerged as a prominent research focus. Vision-language models such as CLIP~\cite{radfordlearning}, DALL-E~\cite{reddydall}, and GPT-4~\cite{OpenAIGPT42023}have demonstrated remarkable performance across a variety of downstream tasks. These large-scale models provide external knowledge for image description, enabling the generation of strong and explicit prompts.
For instance, Zhang \etal~\cite{MultiZhang} proposed an interactive model that facilitates the understanding of video-sentence queries and captures semantic correspondences. Shang \etal~\cite{shang2024prompt} achieved more comprehensive vision-language interactions and fine-grained text-to-pixel alignment through bidirectional prompting. Furthermore, Wang \etal~\cite{CRIS} introduced a CLIP-Driven Referring Image Segmentation framework (CRIS), which effectively transfers multi-modal knowledge to achieve text-to-pixel alignment.
In this paper, we use clip text encoder as a teacher model to \textit{extract semantic embeddings of class names and transfer cross-modal knowledge to the segmentation model to improve the performance} of the model.

\begin{figure*}
     \vspace{-1pt}
     \includegraphics[width=\textwidth]{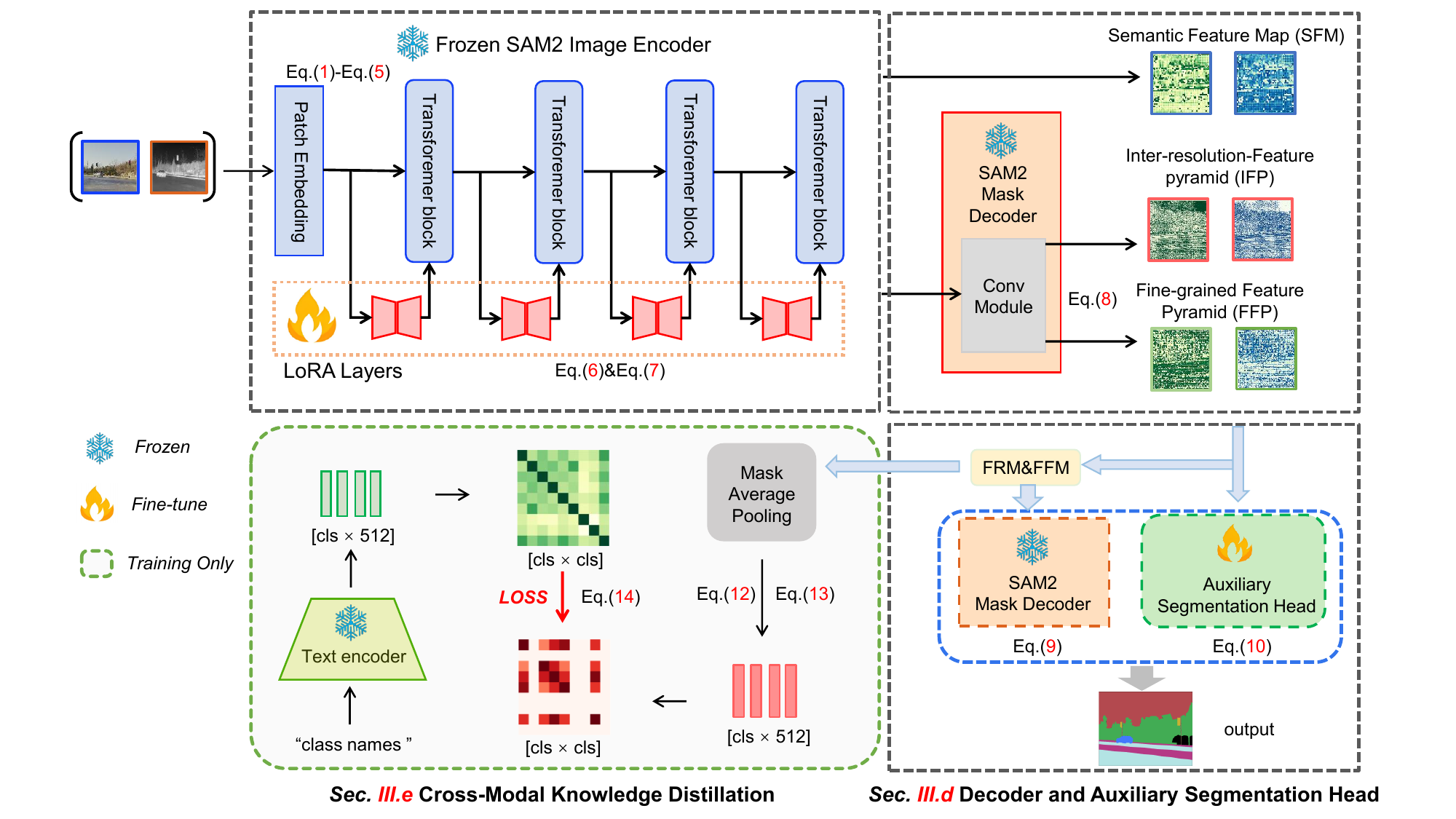}
     \centering
     \vspace{-8pt}
         \caption{Illustration of the proposed \textbf{SARTM} framework for multi-modal semantic segmentation. The architecture combines multi-scale features from a frozen image encoder fine-tuned with LoRA layers. }\label{fig:framework}
\end{figure*}

\section{Method}
\subsection{Framework Overview}
Building upon the SAM2 framework, we introduce a customized SAM2 architecture, referred to as the \text{SARTM} framework, which is specifically optimized for RGB-T semantic segmentation tasks, as shown in Figure~\ref{fig:framework}. The customization starts by freezing the pre-trained image encoder and fine-tuning it through a LoRA layer. This strategy allows the encoder to adapt to the new visual modality while retaining the valuable knowledge learned during the pre-training phase. The image encoder processes the input visual modality \( I \) and produces a semantic feature map (SFM) \( F_n^m \). These feature maps are then passed through the convolutional module of the mask decoder to generate two supplementary feature pyramids: the fine-grained feature pyramid (FFP) \( F_0^m \) and the intermediate-resolution feature pyramid (IFP) \( F_1^m \). Combined with the SFM, these pyramids further enhance the model’s spatial and semantic representation capacity.

To facilitate effective cross-modal feature fusion, we propose a framework that performs a weighted average of the cross-modal representations (SFM, FFP, and IFP), resulting in the feature representation \( \overline{F}_i \), where \( i \in \{0, 1, n\} \). To further enhance the quality of image fusion, we introduce a language-guided fusion module, which leverages language information to guide the feature fusion process and generate the weighted feature maps \( \hat{F}_i \). These fused feature maps are subsequently combined into a unified feature representation \( \tilde{F}_i \), which serves as the input for downstream semantic segmentation tasks.

For the segmentation task, we adopt a dual-path prediction strategy to improve segmentation performance. The first path processes the fused features by feeding them into the SAM2 mask decoder, which employs the frozen Transformer block to derive mask labels from the SFM. These labels then interact with the fine-grained and intermediate-resolution pyramid features, facilitating the construction of high-resolution feature representations. These enhanced high-resolution representations are subsequently processed by the hypernetwork to generate precise segmentation masks, denoted as \( \mathbf{\tilde{S}_0} \).

In the second path, the fusion features are passed to the auxiliary segmentation head, which uses bilinear interpolation to upsample the backbone features, and fuses them with the medium resolution feature pyramid pixel by pixel, then upsamples to the fine-grained pyramid size for fusion, and finally the merged features are processed by the 3×3 convolution layer to enhance the context information, and then upsamples to the target resolution, denoted as \( \mathbf{\tilde{S}_1} \).

In addition, we extract the semantic embedding of class names through CLIP's text encoder as a teacher model to guide the training of segmentation models. In order to realize cross-modal knowledge transfer, we propose an implicit relational transfer mechanism, which distills knowledge by aligning self-similar matrices of image features and text features. By calculating the similarity difference between the student model and the teacher model, the student model can distill the cross-modal semantic knowledge from the teacher model and improve the model performance.

\subsection{LoRA Layers in SAM2's image encoder}
We consider a multi-modal input consisting of \(M\) aligned modalities,
\(
I=\{I^m \in \mathbb{R}^{H\times W\times C}\mid m=1,\dots,M\},
\)
where \(H\), \(W\), and \(C\) denote the spatial resolution and channel dimension, and \(m\) indexes a specific modality (e.g., RGB or thermal). Each modality is fed into the hierarchical Hiera backbone in SAM2 to produce a pyramid of multi-scale representations.

Given an input modality \(I^m\), the patch embedding projects it into a \(d\)-dimensional latent space:
\begin{equation} \label{equ:patch}
    P(I^m) = I^m W_e + b_e,
\end{equation}
where \(W_e \in \mathbb{R}^{C \times d}\) and \(b_e \in \mathbb{R}^{d}\). The resulting feature map is downsampled to \(H_0 = H/s_0\) and \(W_0 = W/s_0\).

As the backbone proceeds through \(n\) hierarchical stages, it gradually decreases spatial resolution while increasing channel capacity, yielding
\begin{equation} \label{equ:msfeature}
    \{I_i^m \in \mathbb{R}^{C_i \times H_i \times W_i} \mid i \in [0, n], \, m \in [1, M]\},
\end{equation}
with
\begin{equation}
    H_i = H / s_i, \quad W_i = W / s_i, \quad s_i = 2^{i+2}.
\end{equation}
Within each stage, features are updated by window-based multi-head self-attention:
\begin{equation} \label{equ:qkv}
    \text{Attention}(Q, K, V) = \text{softmax}\left(\frac{Q K^\top}{\sqrt{d_k}}\right) V,
\end{equation}
where \(Q\), \(K\), and \(V\) are the query, key, and value projections and \(d_k\) is the key dimension.

To enable parameter-efficient, modality-aware adaptation, we augment the query and value projections with LoRA updates. Specifically, we parameterize the residuals with rank-\(r\) factors (\(r \ll d\)):
\begin{equation}\label{equ:qv}
    \Delta Q^m = W_a^Q W_b^Q, \quad \Delta V^m = W_a^V W_b^V,
\end{equation}
where \(W_a^Q, W_a^V \in \mathbb{R}^{d \times r}\) and \(W_b^Q, W_b^V \in \mathbb{R}^{r \times d}\). The resulting augmented projections are
\begin{equation}\label{equ:augmented_projections}
    Q'^m = Q^m + \Delta Q^m, \quad V'^m = V^m + \Delta V^m.
\end{equation}
The LoRA parameters are instantiated per modality and optimized independently, while all backbone weights are kept frozen, yielding efficient cross-modal specialization with minimal additional parameters.

\subsection{Feature Pyramid Network}
We further enhance the hierarchical representations with a Feature Pyramid Network (FPN) that couples lateral connections with a top-down pathway to strengthen multi-scale semantics. At stage \(i\), the modality-specific feature map \(I_i^m\) is first projected by a lateral convolution into a unified embedding space,
yielding \(Z_i^m \in \mathbb{R}^{d \times H_i \times W_i}\). This mapping standardizes the channel dimension to \(d\) while preserving the spatial resolution \((H_i, W_i)\), thereby facilitating subsequent cross-scale fusion within the pyramid.

Let \(\mathcal{L}\) denote the set of stages where top-down aggregation is performed. For each \(i \in \mathcal{L}\), we merge the lateral feature \(Z_i^m\) with the upsampled representation propagated from the next deeper stage, producing the fused pyramid feature \(F_i^m\):
\begin{equation}\label{equ:fusedfm}
  F_i^m =
\begin{cases}
\frac{Z_i^m + \text{Upsample}(F_{i+1}^m)}{2}, & i \in \mathcal{L},\\
Z_i^m, & i \notin \mathcal{L}.
\end{cases}
\end{equation}
Here, \(\text{Upsample}(\cdot)\) resizes \(F_{i+1}^m\) to match the spatial dimensions of \(Z_i^m\), enabling aligned element-wise fusion. This iterative refinement yields a scale-consistent pyramid in which high-level semantics are progressively injected into higher-resolution maps.

The FPN outputs three modality-specific feature maps that capture complementary levels of abstraction and detail: a low-resolution semantic feature map (SFM) \(F_n^m \in \mathbb{R}^{d \times H_n \times W_n}\), and two higher-resolution maps, namely the fine feature pyramid (FFP) \(F_f^m \in \mathbb{R}^{d \times H_0 \times W_0}\) and the intermediate feature pyramid (IFP) \(F_i^m \in \mathbb{R}^{d \times H_1 \times W_1}\).
To improve efficiency and better balance capacity across scales, we compress the channel dimensions of the higher-resolution outputs with \(1\times1\) convolutions, resulting in
\(F_f^m \in \mathbb{R}^{\frac{d}{8} \times H_0 \times W_0}\) and \(F_i^m \in \mathbb{R}^{\frac{d}{4} \times H_1 \times W_1}\), while maintaining their spatial resolutions for subsequent fusion.

\subsection{Decoder and Auxiliary Segmentation Head}
Given the unified fused representation \(\tilde{\mathbf{F}}\), we adopt a dual-branch mask prediction design to produce high-resolution semantic masks. In particular, the semantic feature map \(F_n^m \in \mathbb{R}^{d \times H_n \times W_n}\) is further rectified and fused with the RGB stream by the cross-modal Feature Rectification Module (FRM) and Feature Fusion Module (FFM), resulting in fused features denoted as \(\tilde{\mathbf{f}}\).

\begin{figure*}
      \begin{center}
         \includegraphics[width=0.95\linewidth]{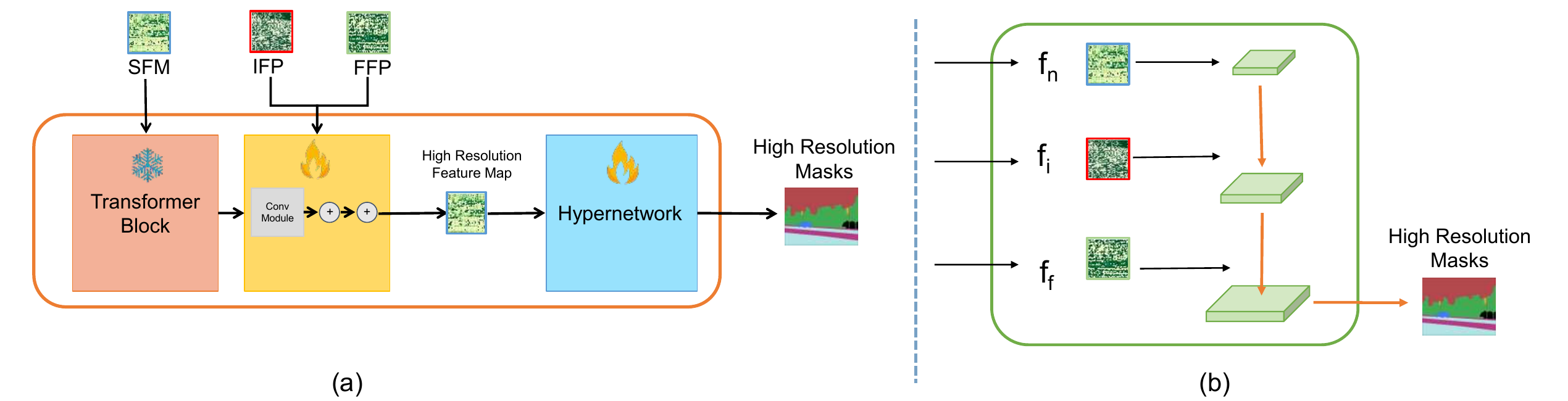}
            \caption{(a) Hierarchical feature fusion via Semantic Feature Map (SFM), Fine-Grained Feature Pyramid (FFP), and Intermediate-Resolution Feature Pyramid (IFP) for cross-scale information aggregation and high-resolution feature map generation.(b) \(f_n\), \(f_i\), and \(f_f\) represent the results of SFM, IFP, and FFP after passing through the FFM and FRM fusion modules, respectively..}\label{fig:decoder}
         \end{center}
    \end{figure*}

\paragraph{Branch I: SAM2 decoder for high-resolution multimasks}
As illustrated in Fig.~\ref{fig:decoder}(a), we extend the SAM2 mask decoder to output high-resolution multimask logits. Let \(\mathcal{C}\) denote the number of semantic categories and \(\mathbf{\tilde{S}}_0 \in \mathbb{R}^{\mathcal{C} \times H_0 \times W_0}\) be the resulting prediction.
We first feed the low-resolution semantic feature \(\tilde{\mathbf{F}}_n \in \mathbb{R}^{d \times H_n \times W_n}\) into a transformer-based decoder \(f_{\text{dec}}\) to obtain coarse logits. These logits are then progressively refined by injecting spatial details from the intermediate pyramid feature \(\tilde{\mathbf{F}}_i \in \mathbb{R}^{\frac{d}{4} \times H_1 \times W_1}\) and the fine-grained feature \(\tilde{\mathbf{F}}_f \in \mathbb{R}^{\frac{d}{8} \times H_0 \times W_0}\). Concretely, we employ bilinear upsampling for resolution alignment and \(1\times1\) convolutions for channel matching:
\begin{equation}\label{eq:combined_high_res_refined}
  \begin{aligned}
      \mathbf{S}_{\text{low}} &= f_{\text{dec}}(\tilde{\mathbf{F}}_n), \\
      \mathbf{S}_{\text{inter}} &= \text{Upsample}(\mathbf{S}_{\text{low}}) + \text{Conv}(\tilde{\mathbf{F}}_i), \\
      \mathbf{\tilde{S}}_0 &= \text{Upsample}(\mathbf{S}_{\text{inter}}) + \text{Conv}(\tilde{\mathbf{F}}_f).
  \end{aligned}
\end{equation}

\paragraph{Branch II: auxiliary FPN-style segmentation head}
In parallel, Fig.~\ref{fig:decoder}(b) depicts an auxiliary head that aggregates multi-scale fused features into a single high-resolution embedding and produces an additional prediction \(\mathbf{\tilde{S}}_1 \in \mathbb{R}^{\mathcal{C} \times H_0 \times W_0}\).
Specifically, we upsample the fused semantic feature \(\tilde{\mathbf{f}}_n \in \mathbb{R}^{d \times H_n \times W_n}\) to match the spatial size of the intermediate feature \(\tilde{\mathbf{f}}_i \in \mathbb{R}^{\frac{d}{4} \times H_1 \times W_1}\), and combine them via element-wise addition. The merged representation is then upsampled to the finest scale and fused with the fine-grained feature \(\tilde{\mathbf{f}}_f \in \mathbb{R}^{\frac{d}{8} \times H_0 \times W_0}\) through additive integration. Finally, a \(3\times3\) convolution enhances local context before generating the auxiliary logits:
\begin{equation}\label{eq:auxiliary_seghead}
\mathbf{\tilde{S}}_1 =
\text{Conv}\!\left(
\text{Upsample}\!\left(\tilde{\mathbf{f}}_i + \text{Upsample}(\tilde{\mathbf{f}}_n)\right)
+ \tilde{\mathbf{f}}_f
\right).
\end{equation}
This dual-branch design provides complementary supervision and improves robustness by explicitly encouraging cross-scale consistency.

\paragraph{Training objective}
We optimize a composite objective that couples cross-entropy supervision with multi-scale predictions. Given pixel-wise labels \(\mathbf{L} \in \mathbb{R}^{H_t \times W_t}\) where \(\mathbf{L}(i,j) \in \{0,1,\dots,\mathcal{C}-1, 255\}\) (with \(255\) indicating ignored regions), we employ an online hard example mining (OHEM) variant of the pixel-wise cross-entropy:
\begin{equation}\label{equ:ohem}
  \mathcal{L}_{\text{CE}}(\mathbf{\tilde{S}}, \mathbf{L}) =
  \frac{1}{n_{\text{min}}}
  \sum_{p \in \mathcal{H}}
  \mathcal{L}_{\text{CE}}(\mathbf{\tilde{S}}(p), \mathbf{L}(p)),
\end{equation}
where \(\mathcal{H}\) denotes the set of hardest pixels selected according to prediction difficulty, and
\(n_{\text{min}} = \max(|\mathcal{H}|, n_{\text{threshold}})\) enforces a minimum number of selected samples with \(n_{\text{threshold}} = n_{\text{total}}/16\), where \(n_{\text{total}}\) is the number of valid pixels.

\subsection{Cross-Modal Knowledge  Distillation}
CLIP~\cite{radfordlearning} optimizes the distance between image and text features in a shared embedding space, aiming to minimize the distance between matched image-text pairs for modality gap mitigation .

As illustrated in Fig~\ref{fig:framework}, we employ CLIP text encoder to extract semantic embeddings  \(T_e\) from category names (e.g., ["Person", ...]) as supervision signals for the segmentation model

The category representations are computed through label \(y \in Y\) and segmentation head outputs using Mask Average Pooling (MAP). 
\begin{equation}
\{f_0, \ldots, f_K\} = MAP(\tilde{\mathbf{f}}_n, y).
\end{equation}

\begin{equation}
\mathcal{L}_{cr} = KL(\tilde{\mathbf{f}}_n, y).
\label{eq:l_se}
\end{equation}

The MAP procedure comprises two key operations: Class-specific feature extraction using y as class masks, followed by average pooling over spatial dimensions to obtain final semantic representations\(\{\tilde{\mathbf{f}}_0, \ldots, \tilde{\mathbf{f}}_K\}\).
To enable cross-modal knowledge transfer, we propose an implicit relation transfer mechanism that distills intra-modality semantic knowledge by aligning the self-similarity matrices of  \(T_e\) and\(\{\tilde{\mathbf{f}}_0, \ldots, \tilde{\mathbf{f}}_K\}\).

\begin{equation}
\mathcal{L}_{se} = KL({Cos}(\{\tilde{\mathbf{f}}_0, \ldots, \tilde{\mathbf{f}}_K\}, \{\tilde{\mathbf{f}}_0, \ldots, \tilde{\mathbf{f}}_K\}^T), {Cos}(T_e, T_e^T)).
\label{eq:l_se}
\end{equation}

where KL denotes the Kullback-Leibler Divergence. After minimizing modality gaps among visual modalities, we strive to reduce the semantic ambiguity by transferring the intra-modal semantic knowledge.

\subsection{Overall Training Objectives}
The overall loss function incorporates the OHEM CrossEntropy loss applied to both \( \mathbf{\tilde{S}_0} \) and \( \mathbf{\tilde{S}_1} \), as defined in Eq.~(\ref{equ:loss}).

\begin{equation}\label{equ:loss}
  \mathcal{L} = w_0 \cdot \mathcal{L}_{\text{CE}}(\mathbf{\tilde{S}_0}, \mathbf{L}) + w_1 \cdot \mathcal{L}_{\text{CE}}(\mathbf{\tilde{S}_1}, \mathbf{L}) + w_2 \cdot \mathcal{L}_{cr} + w_3 \cdot \mathcal{L}_{se}
\end{equation}

where \( w_0, w_1, w_2, w_3 > 0\) are scalar weights that control the relative importance of each loss term.

\begin{figure*}
  \begin{center}
     \includegraphics[width=\linewidth]{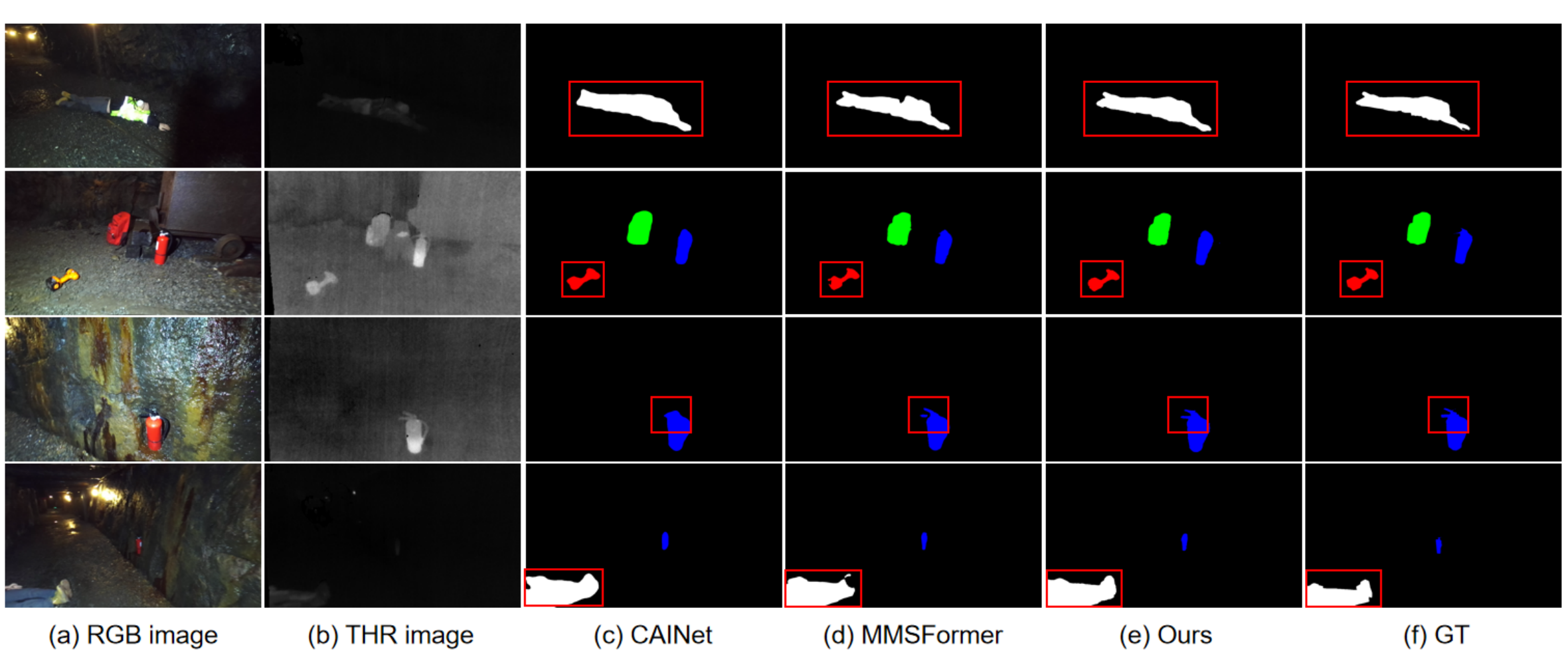}
     \vspace{-25pt}
     \caption{Qualitative results on the PST900 dataset;improvements shown in red.}\label{fig:pst}
  \end{center}
\end{figure*}

\begin{table}[t!]
  \centering
  \caption{Semantic segmentation performance comparison on PST900.}
  \label{tab.pst900}
    \resizebox{\linewidth}{!}{
    \begin{tabular}{llccccc|c}
    \toprule
    \multirow{1}{*}{Methods} & \multirow{1}{*}{Reference} & \multicolumn{1}{c}{\rotatebox{45}{Background}} & \multicolumn{1}{c}{\rotatebox{45}{Fire-Extinguisher}} & \multicolumn{1}{c}{\rotatebox{45}{Backpack}} & \multicolumn{1}{c}{\rotatebox{45}{Hand-Drill}} & \multicolumn{1}{c}{\rotatebox{45}{Survivor}} & \multirow{1}{*}{mIoU} \\
    \midrule
    MFNet \cite{MFNet} & 2017 IROS & 98.6 & 41.1 & 64.2 & 60.3 & 20.7 & 57.0 \\
    RTFNet \cite{sun2019rtfnet}& 2019 RAL & 98.9 & 36.4 & 75.3 & 52.0 & 25.3 & 57.6 \\
    GMNet \cite{zhouGMNet} & 2021 TIP & 99.4 & 85.1 & 83.8 & 73.7 & 78.3 & 84.1 \\
    ABMDRNet \cite{zhang2021abmdrnet} & 2021 CVPR & 98.7 & 24.1 & 72.9 & 54.9 & 57.6 & 67.3 \\
    FEANet \cite{deng2021feanet} & 2021 IROS & - & - & - & - & - & 85.5 \\
    EGFNet \cite{zhou2022edge} & 2022 AAAI & 99.2 & 74.3 & 83.0 & 71.2 & 64.6 & 78.5 \\
    EAEFNet \cite{liang2023explicit} & 2023 RAL & 99.5 & 80.4 & 87.7 & 83.9 & 75.6 & 85.4 \\
    DBCNet \cite{zhoudbcnet} & 2023 TSCS & 98.9 & 62.3 & 71.1 & 52.4 & 40.6 & 74.5 \\
    CAINet \cite{lv2024context} & 2024 TMM & 99.5 & 80.3 & 88.0 & 77.2 & 78.6 &84.7\\
    MRFS\cite{zhangMRFS} & 2024 CVPR & 99.6  & 81.5 & 89.8 & 79.6 & 76.7 & 87.5 \\
    HAPNet \cite{HAPNet} & 2024 CVPR & 99.6 & 81.3 & \textbf{92.0} & \textbf{89.3} & 82.4 & 89.0 \\
    MDRNet \cite{MDRNet} & 2024 TCSVT & 99.07 & 76.75 & 89.41 & 69.46 & 80.17 & 82.97 \\
    RACM-CCSM \cite{peng2024region} & 2024 PR & - & 79.82 & 89.24 & 83.66 & 77.11 & 86.16\\
    AGFNet \cite{AGFNet} & 2025 TITS & 99.50 & 80.30 & 85.30 & 80.20 & 78.7 & 84.8 \\
    MiLNet \cite{MiLNet} & 2025 TIP & 99.54 & 76.25 & 89.64 & 82.15 & 77.91 & 85.11\\
    \midrule
    \textbf{SARTM (Ours)} & - & \textbf{99.7} & \textbf{88.92} & \textbf{92.97} & 83.6 & \textbf{84.19} & \textbf{89.88} \\
    \bottomrule
  \end{tabular}}
  
\end{table}


\section{Experiments}
\subsection{Datasets}
\textbf{Datasets.}
\looseness=-1
To verify the effectiveness of SARTM, we conduct extensive experiments on three publicly available RGB-Thermal (RGB-T) semantic segmentation datasets, namely MFNet~\cite{MFNet}, FMB~\cite{FMB}, PST900~\cite{PST900}.

The details of these datasets are as follows:
\begin{itemize}
    \item \textbf{MFNet dataset} consists of 820 daytime and 749 nighttime RGB-T images, each with a resolution of 640×480. This dataset covers eight common object classes typically encountered in driving scenarios. 
    \item \textbf{PST900 dataset}
    provides a total of 597 and 288 calibrated RGB-T images with a resolution of 1280×720 for training and validation, respectively. Collected from the DARPA Subterranean Challenge, the dataset is annotated with four object classes.
    \item \textbf{FMB dataset}
    contains 1500 image pairs, each consisting of infrared and visible images, annotated with 15 pixel-level categories. The training set comprises 1220 pairs, while the test set includes 280 pairs. 
\end{itemize}

\noindent \textbf{Evaluation Metrics.}
We utilize two commonly adopted evaluation metrics, accuracy (Acc) and intersection over union (IoU), to assess the scene parsing performance for each category.Additionally, the mean values of these metrics across all categories, referred to as mean IoU (mIoU) and mean accuracy (mAcc), are calculated to provide an overall assessment of the network's performance.

\noindent \textbf{Training Settings.}
We enhance our dataset for data augmentation by randomly applying color changes (jittering), mirroring (horizontal flipping), and resizing with extraction (cropping) to both the visible-light (RGB) and infrared (thermal) images. We use the AdamW optimizer~\cite{loshchilov2017adamw} with an initial learning rate 1e-4 and weight decay 0.01.
The model is trained with a batch size of 8 for 500 epochs.
We use cross-entropy loss function.

\begin{table*}[!t] 
  \centering
  \caption{
Semantic segmentation performance comparison on MFNet.}
  \label{tab.mfnet}
  \renewcommand{\tabcolsep}{1pt}
  \resizebox{\linewidth}{!}{
\begin{tabular}{llcccccccc|c}
      \midrule
    \multirow{1}{*}{Methods} & \multirow{1}{*}{Pub.} & \multicolumn{1}{c}{Unlabeled} & \multicolumn{1}{c}{Car} & \multicolumn{1}{c}{Person} & \multicolumn{1}{c}{Bike} & \multicolumn{1}{c}{Curve} & \multicolumn{1}{c}{Car Stop}  &  \multicolumn{1}{c}{Color Cone} & \multicolumn{1}{c}{Bump} & \multirow{1}{*}{mIoU} \\
    \midrule
    MFNet~\cite{MFNet} & 2017 IROS & 96.9 & 65.9 & 58.9 & 42.9 & 29.9 & 9.9   &25.2 & 27.7 & 39.7 \\
    RTFNet ~\cite{sun2019rtfnet} & 2019 RAL & 98.5 & 87.4 & 70.3 & 62.7 & 45.3 & 29.8   & 29.1 & 55.7 & 53.2 \\
    FuseSeg ~\cite{sun2020fuseseg} & 2020 TASE    & 97.6 & 87.9 & 71.7 & 64.6 & 44.8 & 22.7 &  46.9 & 47.9  & 54.5 \\
    ABMDRNet ~\cite{zhang2021abmdrnet} & 2021 CVPR & \textbf{98.6} & 84.8 & 69.6 & 60.3 & 45.1 & 33.1 &47.4 & 50.0  & 54.8 \\
    FEANet ~\cite{deng2021feanet}& 2021 IROS & 98.3 & 87.8 & 71.1 & 61.1 & 46.5 & 22.1  &55.3 & 48.9  & 55.3 \\
    EGFNet ~\cite{zhou2022edge} & 2022 AAAI & - & 87.6 & 69.8 & 58.8 & 42.8 & 33.8 &48.3 & 47.1  & 54.8 \\
    SegMiF+ ~\cite{liu2023segmif}  & 2023 ICCV & 98.1 & 87.8 & 71.4 & 63.2 & 47.5 & 31.1  &48.9 & 50.3 & 56.1  \\
    MDRNet ~\cite{MDRNet}  & 2023 TNNLS & -& 87.1 & 69.8 & 60.9 & 47.8 & 34.2   &50.2 & 55.0 & 56.8  \\
    EAEFNet \cite{liang2023explicit} & 2023 RAL & -& 87.6 & 72.6 & 63.8 & 48.6 & 35.0  &52.4 & 58.3  & 58.9 \\
    CAINet ~\cite{lv2024context} & 2024 TMM & -& 88.5 & 66.3 & \textbf{68.7} & \textbf{55.4} & 31.5  &48.9 & 60.7 & 58.6 \\
    CMX ~\cite{CMX} & 2024 TITS & 98.3 & 90.1 & \textbf{75.2} & 64.5 & 50.2 & 35.3   &\textbf{54.2} & 60.6 & 59.7 \\
    RACM-CCSM ~\cite{CMX} & 2024 PR & - & 89.8 & 75.1 & 65.4 &45.7 & 37.4   &56.1 & 61.6 & 60.0 \\
    LLESeg ~\cite{LLESeg} & 2025 TIM & 98.3 & 88.5 & 73.2 & 64.8 & 46.8 & 30.0 & 52.5 & 62.4 & 58.4 \\
    AGFNet ~\cite{AGFNet} & 2025 TITS & - & 88.0 & 73.7 & 64.6 & 45.3 & 34.3  &  57.0 & 52.5 & 58.6 \\
    \midrule
    \textbf{SARTM (Ours)} & - & 98.4 & \textbf{91.32} & \textbf{75.42} & 65.22 & 49.97 & \textbf{49.47} & \textbf{54.96} & 55.53  & \textbf{60.03}\\
    \midrule
  \end{tabular}
  }
\end{table*}

\begin{figure}
  \begin{center}
     \includegraphics[width=\linewidth]{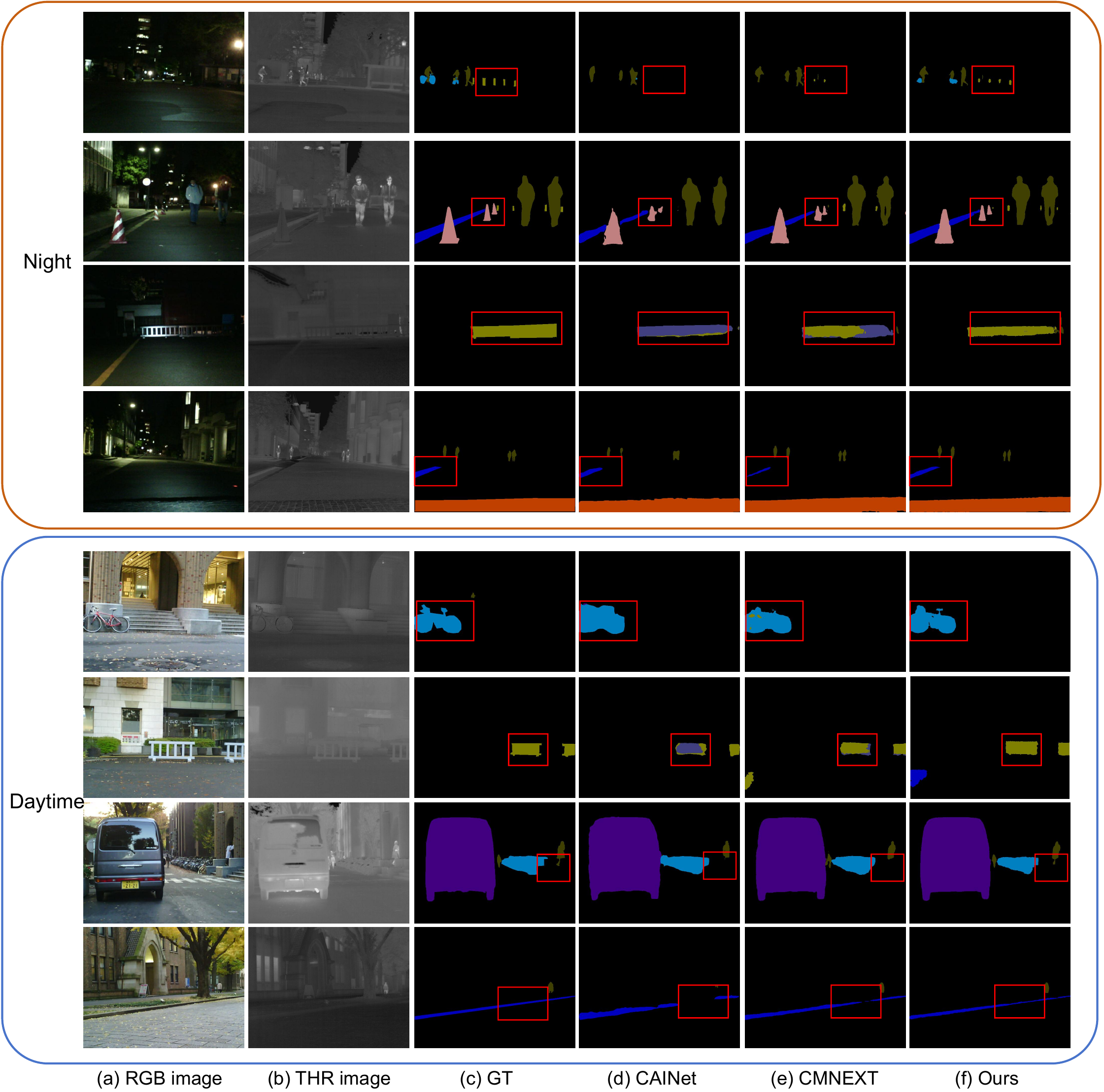}
     \caption{Qualitative comparisons on the MFNet dataset. We can see that our SARTM can provide better results in various lighting conditions and environments. The comparison results demonstrate our superiority.}\label{fig:mf}
  \end{center}
\end{figure}

\subsection{Comparisons with SoTA Scene Parsing Networks}
We perform quantitative comparisons with 15, 14, and 12 state-of-the-art (SoTA) RGB-T scene parsing networks on the PST900~\cite{PST900}, MFNet ~\cite{MFNet}, and FMB~\cite{FMB} datasets, respectively. The results are presented in Tables~\ref{tab.pst900}, ~\ref{tab.mfnet}, and ~\ref{tab.fmb}. Furthermore, we provide qualitative comparisons on these three datasets, as illustrated in Figures~\ref{fig:pst}, Figures~\ref{fig:mf}.

The experimental results of our proposed SARTM on the PST900 underground dataset are shown in Table~\ref{tab.pst900}, demonstrating the superior performance of our framework in key safety perception for underground environments.  Our method achieves a state-of-the-art mIoU of 89.88\%, representing a 0.88\% absolute improvement over HAPNet~\cite{HAPNet}, with particularly significant advancements in the life-critical categories.

Furthermore, the proposed method is applicable to various datasets, demonstrating strong performance across different scenarios.  Specifically, as shown in Table~\ref{tab.mfnet} and Table~\ref{tab.fmb}, on the MFnet dataset, although our method slightly underperforms compared to the best-performing method in terms of mIoU, the difference is marginal, indicating that our framework achieves near-optimal performance on this dataset.  On the FMB dataset, our method outperforms the comparison methods, further validating the effectiveness of SARTM in diverse datasets.

Therefore, despite minor discrepancies on some datasets, our method shows broad application and robust performance across different environments, making it highly effective for real-world applications.

\subsection{Qualitative Analysis}
In addition to the quantitative analysis, we conducted a qualitative evaluation of the predicted segmentation maps. As shown in Fig~\ref{fig:pst}, we compare prediction results on the PST900~\cite{PST900} dataset with those of CAINet~\cite{lv2024context} and MMSFormer~\cite{Reza2024MMSFormer}. The figure displays the input RGB image, thermal image, ground truth segmentation map and the model's predictions. As highlighted by the rectangular bounding boxes, our model demonstrates superior accuracy in detecting objects with more precise contours compared to the other two metods.

Fig~\ref{fig:mf} presents the segmenation results predicted by CMNext~\cite{CMnext},CAINet~\cite{lv2024context} and our SARTM model. As emphasized within the rectangular borders, our proposed model exhibits strong segmentation performance, particularly for objects such as color cone, fine details of figure, and curve.

\begin{figure*}
  \begin{center}
     \includegraphics[width=\linewidth]{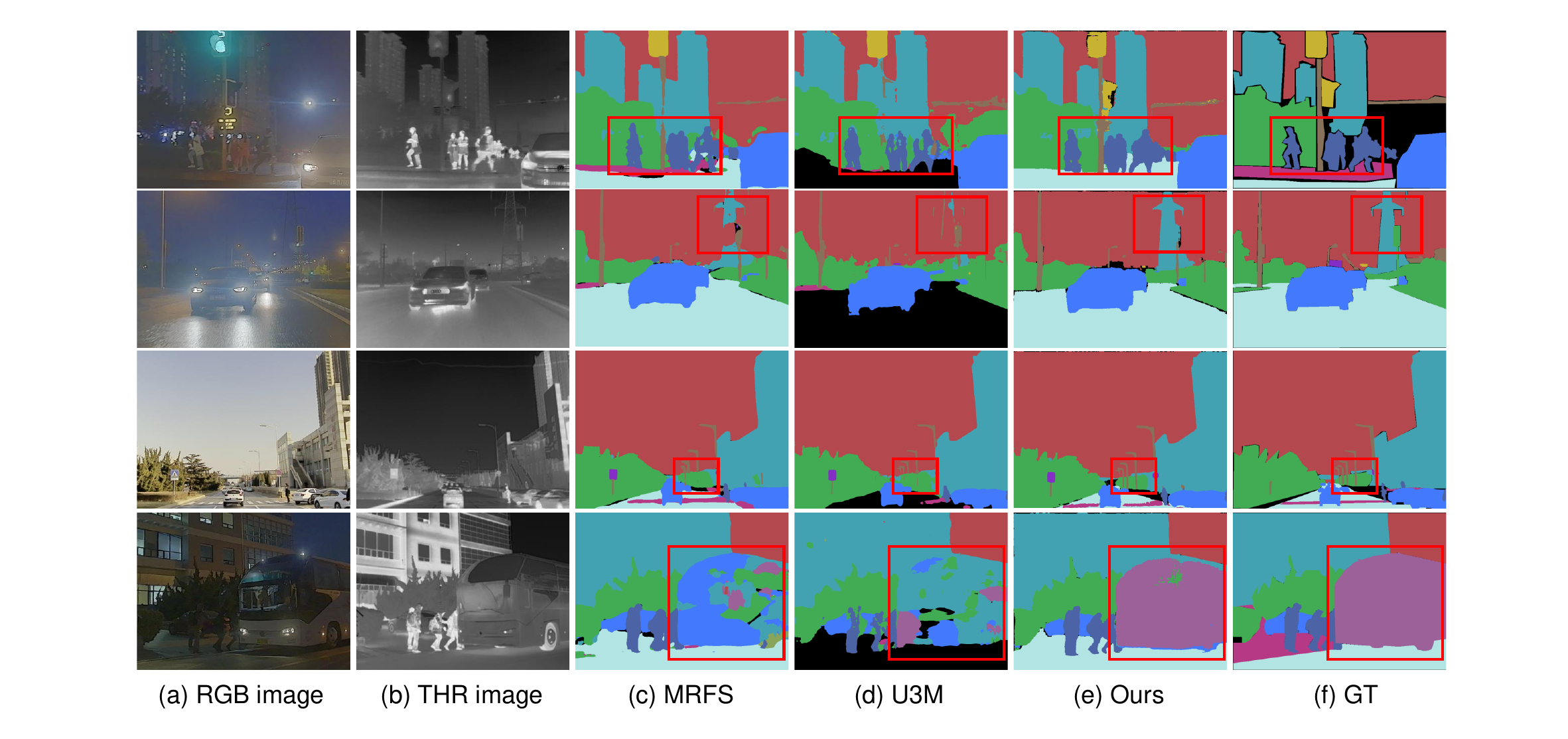}
     \vspace{-25pt}
     \caption{Qualitative results on the FMB dataset; improvements shown in red.}\label{fig:fmb}
  \end{center}
\end{figure*}

Fig~\ref{fig:fmb} illustrates the capability of effectively utilizing thermal data to generate more coherent segmentations. For instance, for people in nighttime conditions, the segmentation performance of the baseline model is poor due to lighting issues.However, our method successfully identifies the person as a single entity, highlighting its remarkable ability to leverage thermal data for segmentation. Compared to other methods, our approach exhibits strong segmentation performance across categories such as bus, person, and pole.

\begin{table}[h!]
        \caption{ Semantic segmentation performance comparison on FMB.}~\label{tab:data_set_results_seg}
        \label{tab.fmb}
  \resizebox{\linewidth}{!}{
        \begin{tabular}{lccccccccc|c}
	\midrule
        \multirow{1}{*}{Methods}& \multirow{1}{*}{Reference} & \multicolumn{1}{c}{Car} & \multicolumn{1}{c}{Person} & \multicolumn{1}{c}{Truck} & \multicolumn{1}{c}{T-Lamp} & \multicolumn{1}{c}{T-Sign} & \multicolumn{1}{c}{Building} & \multicolumn{1}{c}{Vegetation} & \multicolumn{1}{c}{Pole} & \multirow{1}{*}{mIoU} \\
        \midrule
	DIDFuse \cite{zhao2020didfuse}&2020 IJCAI & 77.7 & 64.4 & 28.8 & 29.2 & 64.4 & 78.4 & 82.4 & 41.8 & 50.6 \\
	U2Fusion \cite{U2Fusion2020}& 2020 TPAMI & 76.6 & 61.9 & 14.4 & 28.3 & 68.9 & 78.8 & 82.2 & 42.2 & 47.9 \\
	FEANet \cite{deng2021feanet}&2021 IROS & 73.9 & 60.7 & 32.3 & 13.5 & 55.6 & 79.4 & 81.2 & 36.8 & 46.8 \\  
	GMNet\cite{zhou2021gmnet}& 2021 TIP & 79.3 & 60.1 & 22.2 & 21.6 & 69.0 & 79.1 & 83.8 & 39.8 & 49.2 \\
	LASNet\cite{li2022rgb}& 2022 TCSVT & 72.6 & 48.6 & 14.8 & 2.9 & 59.0 & 75.4 & 81.6 & 36.7 & 42.5 \\
	ReCoNet \cite{reconet} & 2022 ECCV & 75.9 & 65.8 & 14.9 & 34.7 & 66.6 & 79.2 & 81.3 & 44.9 & 50.9 \\
	TarDAL \cite{TarDAL}& 2022 CVPR& 74.2 & 56.0 & 18.8 & 29.6 & 66.5 & 79.1 & 81.7 & 41.9 & 48.1 \\
    SegMiF \cite{liu2023segmif}&2023 ICCV& 78.3 & 65.4 & 47.3 & 43.1 & 74.8 & 82.0 & 85.0 & 49.8 & 54.8 \\
    MRFS\cite{zhangMRFS}& 2024 CVPR & 76.1 & \textbf{71.3} & 34.4 & \textbf{50.0} & 75.8 & \textbf{85.4} & \textbf{86.9} & \textbf{53.64} & 61.1 \\
    U3M \cite{li2024u3m}& 2025 PR& \textbf{82.3} & 65.97 & 41.93 & 46.22 & \textbf{81.0} & 81.3 & 86.76 & 48.76 & 60.76 \\
        \midrule
	 \textbf{SARTM (Ours)}&- & 78.3 & 65.4 & \textbf{47.3} & 43.1 & 74.8 & 82.0 & 85.0 & 49.8 & \textbf{61.57} \\
        \midrule
  \end{tabular}}
\end{table}

\subsection{Ablation Study}
\subsubsection{Impact of Rank Size on Performance in LoRA Layers}
LoRA is utilized because it provides a lightweight method to adapt pre-trained models to our task without retraining all the parameters, which is computationally expensive.
In this study, we investigate the impact of LoRA rank on the performance of SARTM. We tested various LoRA ranks (2, 4, 16, 32, and 64) . Table~\ref{tab:lora} shows the performance of SARTM with different LoRA ranks. Our results indicate that, within a certain range of ranks, the performance of SAM steadily improves. However, when the rank exceeds 16, performance begins to degrade. This trend suggests that returns diminish once the LoRA rank surpasses the optimal point, with Rank 16 yielding the best overall segmentation performance.

\begin{table}[t]
\centering
\begin{minipage}[t]{0.48\linewidth}
  \centering
  \caption{Ablation study on the rank size on the LoRA layer}
  \label{tab:lora}
  \footnotesize
  \begin{tabular}{ccc}
    \midrule
    Rank size & mAcc & mIoU\\
    \midrule
    4  & 92.01 & 87.68 \\
    8  & 92.55 & 88.11 \\
    16 & \textbf{93.71} & \textbf{89.88} \\
    32 & 92.73 & 89.01 \\
    64 & 91.95 & 88.80 \\
    \midrule
  \end{tabular}
\end{minipage}\hfill
\begin{minipage}[t]{0.48\linewidth}
  \centering
  \caption{Performance of various auxiliary segmentation heads on FMB.}
  \label{tab:ablation_seg_block}
  \footnotesize
  \begin{tabular}{lc}
    \midrule
    Auxiliary Segmentation Head & mIoU (\%) \\
    \midrule
    FPN      & 58.85 \\
    DeepLab  & 57.58 \\
    SegFormer& 56.94 \\
    \midrule
  \end{tabular}
\end{minipage}
\end{table}


\subsubsection{Effectiveness of Key Components}
We conducted a series of ablation studies to investigate the contribution of each component within the fusion module to the overall model performance. As shown in Table~\ref{tab:ablation_block}, the results highlight the importance of these components.Initially, In the RGB and Thermal modalities of the FMB dataset,we observed that the absence of language guidance resulted in a 3\% performance degradation, indicating that aligning knowledge across modalities through the calculation of feature similarities significantly aids in effective feature extraction. Furthermore, removing the auxiliary segmentation head led to a 4\% performance drop. The FPN module, renowned for its ability to aggregate contextual information at multiple scales, substantially enhances the model’s capacity to capture and integrate multi-scale contextual information, which is crucial for achieving precise segmentation. Similar effects were observed on the other two datasets, further validating the effectiveness of our approach. These comprehensive ablation studies collectively underscore the importance of each component in the fusion module, revealing that every module plays a unique and vital role in achieving the overall performance of the model.

\begin{table}[t!]
  \centering
  \caption{Ablation study of the different Block on three dataset.}
  \label{tab:ablation_block}
  \renewcommand{\tabcolsep}{2.5pt}
  \begin{tabular}{lcccc}
    \midrule
    \multicolumn{1}{c}{Structure} & \multicolumn{1}{c}{FMB} & \multicolumn{1}{c}{MFNet} & \multicolumn{1}{c}{PST900} \\ 
    \midrule
    SARTM  &  \textbf{61.57}  &  \textbf{60.03} &  \textbf{89.88}\\
    - w/o Aux\_Head   & 58.85 (-2.72) & 59.32 (-0.71) & 86.24 (-3.64) \\
    - w/o CMKD  & 57.12 (-4.4)  & 58.67 (-1.36) & 84.92 (-4.96) \\
    \midrule
  \end{tabular}
\end{table}

\begin{figure*}
  \begin{center}
     \includegraphics[width=0.9\linewidth]{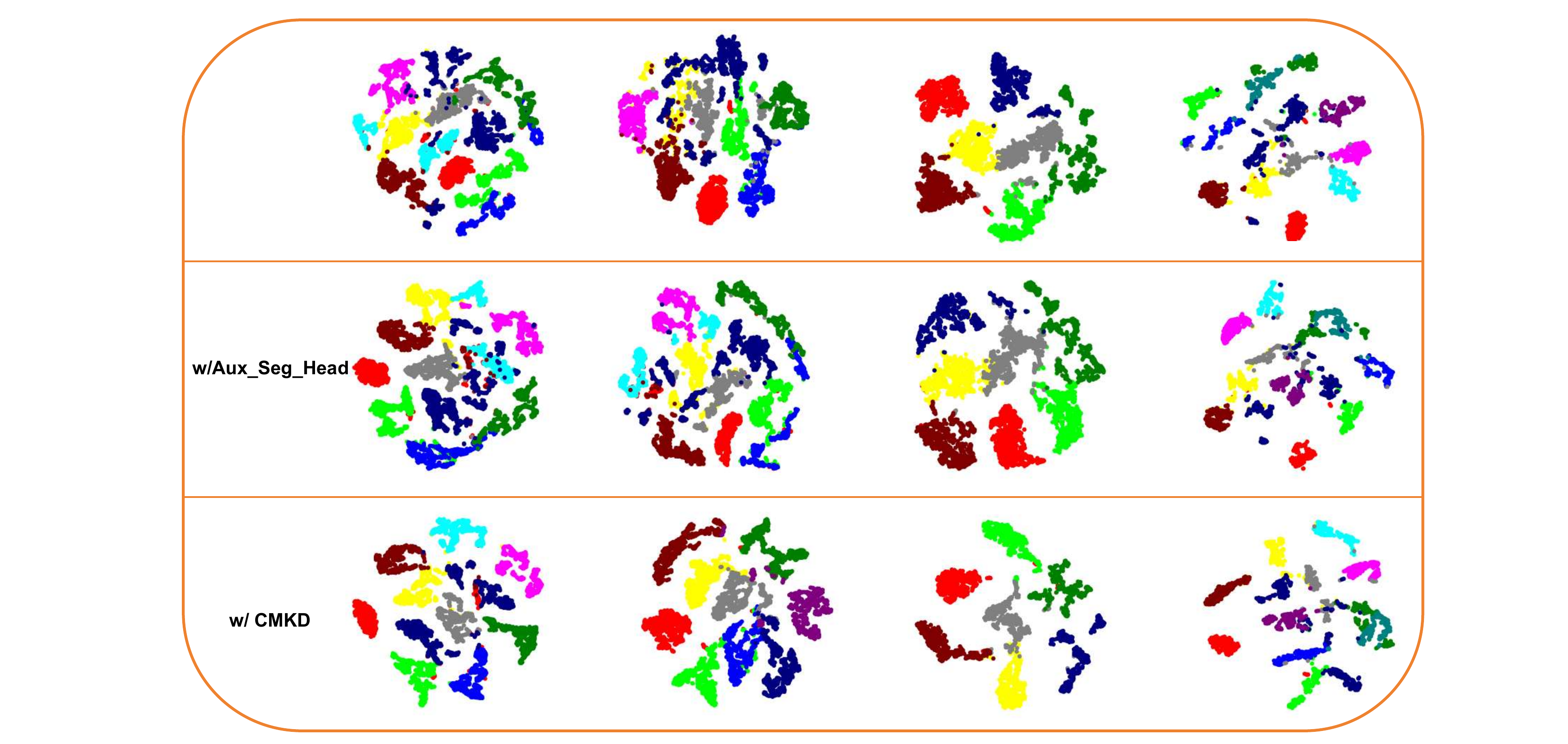}
     \vspace{-10pt}
     \caption{t-SNE visualization on FMB dataset. Each row compares a different segmentation strategy.}\label{fig:tsne}
  \end{center}
\end{figure*}

\begin{figure*}
     \includegraphics[width=\linewidth]{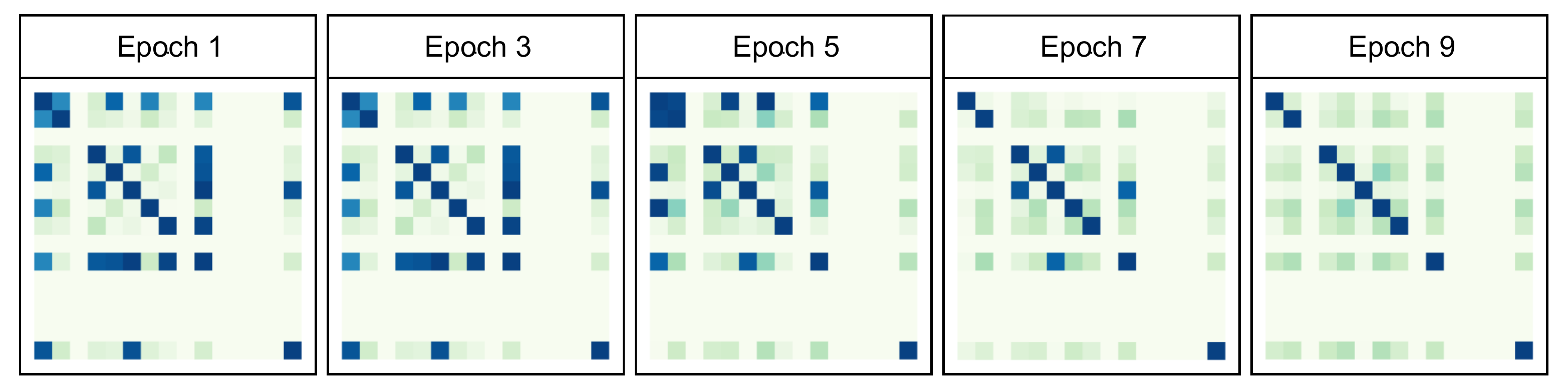}
     \caption{Visualization of the Similarity Matrix Across Different Epochs.}\label{fig:language_vis}
\end{figure*}

\subsubsection{Effectiveness of Loss Functions}
We conducted an ablation study on the proposed loss functions, and the results presented in the table demonstrate that each of the loss functions contributes positively to the improvement of model performance. Specifically, the experimental results highlight the distinct contributions of each loss function to the overall performance. Initially, without any additional loss functions, the model achieved a mean Intersection over Union (mIoU) of 84.95, an F1 score of 90.43, and an accuracy of 87.48. After incorporating the Cross-Entropy (CE) loss, the model's performance improved by 0.84 in mIoU, 0.89 in F1 score, and 0.79 in accuracy, indicating a significant impact of the CE loss on performance enhancement. The best performance was achieved when all loss functions were combined, further validating their effectiveness in optimizing the model.
\begin{table}[!t]
	\centering
    \caption{Ablation of different losses on PST900 with SARTM.}
    \label{tab:rgbt_results}
    \renewcommand{\tabcolsep}{4pt}
    \begin{tabular}{l|c|c|c|ccccccc}
    \midrule
    $\mathcal{L}_{\text{CE}}$ & $\mathcal{L}_{\text{CE}}$ & $\mathcal{L}_{cr}$ & $\mathcal{L}_{se}$ & mIoU & $\Delta$ & F1 & $\Delta$& Acc & $\Delta$ \\
    \midrule
    \checkmark &&&& 84.95 && 90.43 & &87.82 & & \\ \midrule
    \checkmark &\checkmark & &&85.79 & +0.84 & 91.32 & +0.89 & 89.27& +1.45 \\ \midrule
    \checkmark &\checkmark &\checkmark && 87.14 & +1.35 & 92.87 & +1.55 & 91.01 & +1.74 \\ \midrule
    \checkmark&\checkmark & \checkmark & \checkmark & \textbf{89.88} & +1.84 & \textbf{94.57} & +1.70 & \textbf{93.7} & +2.29 \\ \midrule
    \end{tabular}
\end{table}

\subsubsection{Impact of losses weight on Performance in SARTM}
The impact of the weight of different loss functions on the performance of SARTM was evaluated through an ablation study on the MFNet dataset. The results, shown in Table \ref{tab:loss}, indicate that varying the weights of the loss terms significantly influences the model's performance.
For ${w}{1}$, the highest mIOU value of 58.67 was achieved when ${w}{1}$ was set to 0.008, indicating an optimal balance between the loss terms. Increasing ${w}{1}$ to higher values, such as 0.03 or 0.05, led to a decrease in mIOU, with values of 54.53 and 57.81, respectively. This suggests that excessively large or small weights for ${w}{1}$ may negatively impact the model's performance.
For ${w}{2}$, a weight of 10000 achieved the highest mIOU of 59.32, which was slightly lower than the peak performance of 60.03 when ${w}{3}$ was set to 100. These findings highlight the importance of properly tuning the loss weights to maximize performance, as extreme values for any of the loss terms could lead to suboptimal results.
In summary, the optimal configuration for maximizing mIOU is achieved by setting ${w}{1}$ to 0.008, ${w}{2}$ to 10000, and ${w}_{3}$ to 100, resulting in a final mIOU of 60.03, demonstrating the importance of fine-tuning the loss weight parameters in SARTM.

\begin{table}[h!]
        \centering
        \caption{Ablation study of the losses weight on MFNet dataset.}
        \label{tab:loss}
        \begin{tabular}{lc|cc|cc}
        \midrule
        ${w}_{1}$ & mIOU & ${w}_{2}$ & mIOU &  ${w}_{3}$  & mIOU \\
        \midrule
        0.05  & 57.81         & 10      & 57.42       &  1    &58.05     \\
        0.03  & 54.53         & 100    &57.16         &  10   &57.56    \\
        \textbf{0.008} & \textbf{58.67}& 1000 &57.42 & 50   &59.16   \\
        0.004 & 57.46         & \textbf{10000} &\textbf{59.32} &  \textbf{100}  & \textbf{60.03} \\
        0.002 & 56.91         & 100000  &58.64  &  1000 &57.63  \\
        \midrule
\end{tabular}
\end{table}

\subsubsection{Impact of Auxiliary Segmentation Head on Performance}
To further examine the impact of the auxiliary segmentation head on the performance of the fusion block, we compared three different segmentation heads: FPN, DeepLab, and SegFormer. The experimental results, presented in Table~\ref{tab:ablation_seg_block}, show that the FPN segmentation head outperforms the others, achieving an mIoU of 58.85\%. In contrast, the DeepLab segmentation head yields an mIoU of 57.58\%, and the SegFormer segmentation head achieves 56.94\%. These findings indicate that the FPN segmentation head provides superior performance in this experiment, significantly enhancing the overall performance of the fusion block. 

\subsection{Visualization Analysis}
\subsubsection{T-SNE Visualization}
The feature clustering results extracted by SARTM on the FMB dataset, as shown in Figure~\ref{fig:tsne}.
The baseline model's feature distribution (top row) is plagued by high intra-class variance and low inter-class separability. Feature points of the same class are scattered, indicating a failure to learn robust representations against cross-modal visual variations. Furthermore, ambiguous, overlapping clusters manifest the core problem of semantic ambiguity.

Integrating the auxiliary segmentation head (middle row) improves cluster delineation by providing richer contextual information from multi-scale features. However, the persistence of inter-class overlap demonstrates that enhancing visual features alone is insufficient to fully resolve the underlying semantic ambiguities.

The introduction of language guidance (bottom row) fundamentally restructures the feature manifold. It achieves high intra-class compactness by leveraging language as a modality-invariant semantic anchor, forcing disparate visual features into tight, unified clusters. Concurrently, it dramatically enhances inter-class separability via relational knowledge distillation. This mechanism transfers the semantic structure from the language domain to the visual feature space, enabling the model to internalize the relationships between categories.

\subsubsection{Language-aided Distillation Visualization}
We visualized the language module, and as shown in Fig~\ref{fig:language_vis}.
Initially, at Epoch 1, the matrix exhibits a noisy and unstructured pattern. The diagonal is weak, and off-diagonal values are high and diffuse, indicating that the untrained model lacks semantic discrimination. At this stage, the feature space is disorganized, with no clear distinction between intra-class and inter-class similarities.

As training progresses, a clear structural evolution emerges. The diagonal elements, representing intra-class similarity, become progressively stronger, signifying that the model is learning to group instances of the same class. Concurrently, the off-diagonal values are suppressed as the matrix becomes sparser. This transition demonstrates the efficacy of the relational knowledge distillation, as the model begins to learn the semantic distances between categories, guided by the language-based teacher model.

By the final training stages (Epoch 9), the matrix has converged to a highly structured semantic relationship map. It is characterized by a dominant diagonal, indicating robust intra-class consistency, and heavily attenuated off-diagonal regions, which represent well-defined inter-class boundaries. This final state can be interpreted as a successful projection of the semantic relationships from the language space onto the visual feature space.

\subsubsection{Modality Complementarity and Fused-Prototype Alignment}
\begin{figure}
    \begin{center}
     \includegraphics[width=\linewidth]
     {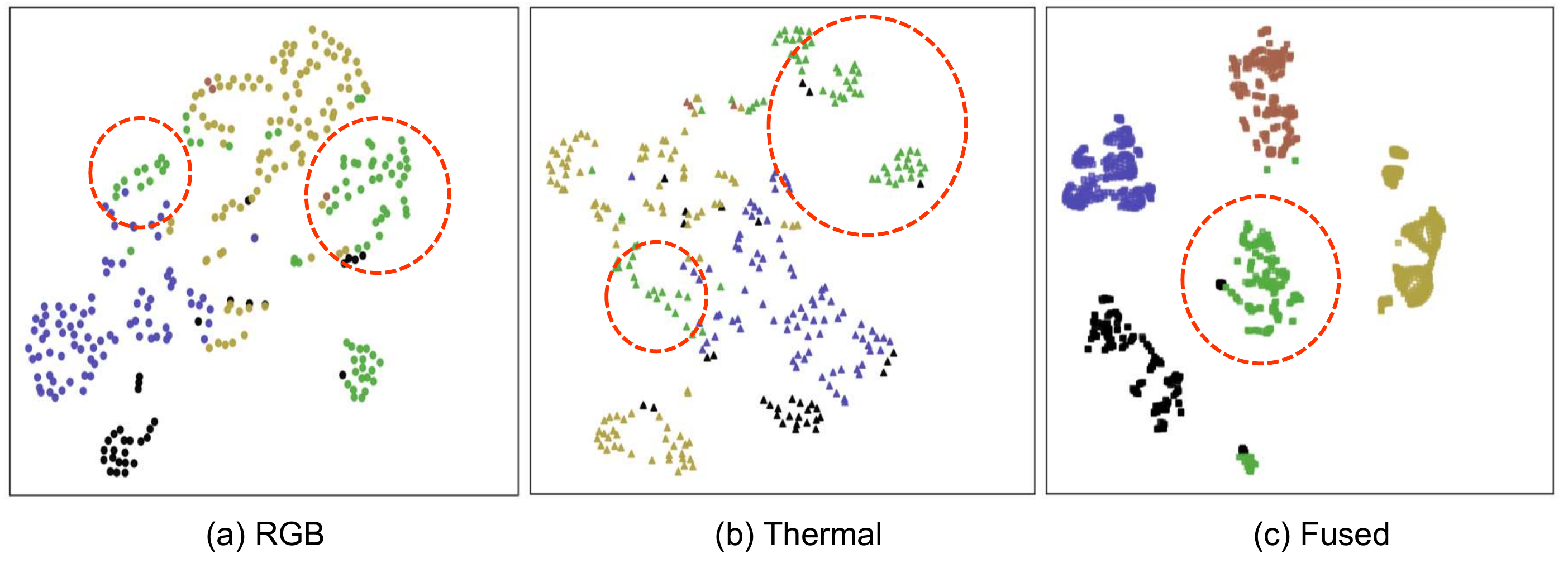}
     \caption{t\mbox{-}SNE visualization of per\mbox{-}pixel embeddings from (a) RGB, (b) thermal, and (c) fused representations.}
\label{fig:Fused}
    \end{center}
\end{figure}

Figure~\ref{fig:Fused}. t-SNE of per-pixel embeddings from RGB (left), thermal (middle), and our fused space (right). RGB features suffer from illumination-induced overlap; thermal features are fragmented due to low texture and sensor non-uniformity. After fusion and prototype alignment, categories form compact, well-separated clusters, evidencing improved intra-class compactness and inter-class margins and validating our “fuse-then-prototype” design.

\begin{figure}[t!]
    \begin{center}
     \includegraphics[width=\linewidth]
     {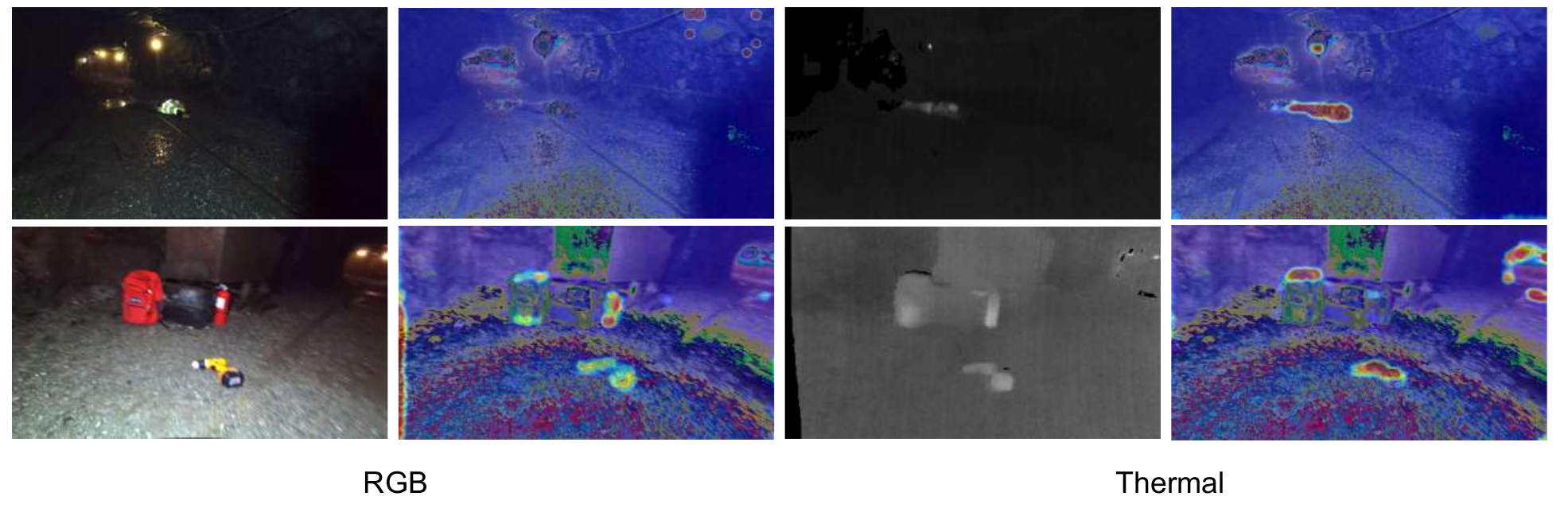}
     \caption{Grad-CAM visualizations on RGB and Thermal.}
\label{fig:CAM}
    \end{center}
\end{figure}

Figure~\ref{fig:CAM}. RGB and Thermal exhibit complementary behaviors: RGB excels at delineating boundaries and geometric structures, whereas Thermal emphasizes semantic cores and heat-emitting regions.   By aligning fine-grained RGB boundaries with the stable thermal cores in the fused representation, the approach effectively suppresses reflective artifacts in RGB and, leveraging Thermal’s reliable responses to true heat sources and human-contact objects, improves overall robustness.
FEANet

\subsection{Model Efficiency}
As summarized in Table~\ref{tab.para}, our method attains favorable trade-offs between model size and inference speed. The proposed modules are realized entirely in low-rank spaces, enabling a markedly smaller parameter footprint while sustaining rapid throughput. Consequently, our model offers significant advantages for practical applications.
\begin{table*}[t!]
    \centering
    \caption{Comparison of model parameters and inference speed. FPS of previous methods was measured under the same environment as ours.}
    \label{tab.para}
    \renewcommand{\tabcolsep}{10pt}
    \resizebox{\linewidth}{!}{%
    \begin{tabular}{ccccc}
    \toprule
    Methods & Input Size & Trainable Params (M)& Total Params (M) & FPS \\
    \midrule    
    EGFNET \cite{zhou2022edge} & $640 \times 480$ & $28.18$  & $28.18$ & $9.63$ \\
    CMX \cite{CMX} & $640 \times 480$ & $181.1$ & $181.1$ & $9.0$ \\ 
    MMSFormer \cite{Reza2024MMSFormer} & $640 \times 480$ & $61.26$ & $61.26$ & $23.41$ \\
    CMNeXt \cite{CMnext} & $640 \times 480$ & $116.5$ & $116.5$  & $12.4$ \\
    \midrule
    \textbf{SARTM (Ours)}  & $640 \times 640$ & \textbf{9.22} & $237.38$ & $13.73$ \\
    \bottomrule
    \end{tabular}}
\end{table*}

\subsection{Comparison with SHIFNet}
We further compare our method with SHIFNet~\cite{zhao2025unveiling}, a recent SAM2-driven RGB-Thermal segmentation framework with language guidance.

\paragraph{Methodological differences.}
SHIFNet is built upon a \emph{hybrid interaction paradigm} that explicitly addresses SAM2's modality bias via two dedicated components: (i) \emph{Semantic-Aware Cross-modal Fusion (SACF)}, which learns text-guided affinity to dynamically reweight RGB/thermal features across multiple pyramid levels; and (ii) a lightweight \emph{Heterogeneous Prompting Decoder (HPD)} that injects category embeddings (from LanguageBind) and global semantic enhancement to improve cross-modal semantic consistency.
In contrast, our SARTM follows a \emph{parameter-efficient adaptation} route: we inject LoRA updates into the query/value projections of the SAM2 image encoder while keeping backbone weights frozen, and employ a dual-path prediction design (SAM2 decoder head plus an auxiliary FPN-style segmentation head). Moreover, instead of learning text-guided fusion weights, we introduce a language-guided \emph{prototype alignment} objective that matches visual class prototypes with text-derived inter-class similarity, providing a complementary semantic regularization.

\paragraph{Performance comparison.}
On PST900, SARTM achieves \(89.88\%\) mIoU, slightly surpassing SHIFNet (\(89.8\%\), \(+\!0.08\%\)). On MFNet, SARTM reaches \(60.03\%\) mIoU, improving over SHIFNet (\(59.2\%\), \(+\!0.83\%\)). These results indicate that our adaptation strategy is particularly competitive on PST900/MFNet.

\paragraph{Efficiency comparison.}
SHIFNet reports \(32.27\)M trainable parameters. Our SARTM requires only \(9.22\)M trainable parameters, yielding a \(3.50\times\) reduction (\(\approx 71.4\%\) fewer trainable parameters). In terms of runtime, SHIFNet reports \(123\) ms/image on an RTX A6000 (approximately \(8.13\) FPS) at \(480\times640\) input, whereas SARTM runs at \(13.73\) FPS with \(640\times640\) input under our evaluation setting, suggesting a favorable speed--accuracy trade-off.\footnote{Note that the reported input resolutions differ; nevertheless, both results consistently reflect the efficiency advantage of low-rank adaptation.}

\section{Conclusion}
This paper presents SARTM, a novel adaptation of the SAM2 architecture specifically tailored for RGB-T semantic segmentation tasks. SARTM integrates LoRA-based adaptability, which enhances the SAM model to better align with the requirements of our task. Furthermore, we introduce a dual-channel prediction mechanism and multi-scale fusion features, which significantly improve segmentation accuracy. Finally, we utilize language as the guiding modality. The experimental results demonstrate that, with the language-aided, the model can more effectively achieve the alignment between the category modality and the feature modality. This allows for a deeper understanding of the context and enhances the model's capability in category discrimination, enabling the model to identify and process information of different categories more accurately, thereby improving the model's performance. Extensive experiments demonstrate that SARTM exhibits superior performance on RGB-T semantic segmentation benchmarks.

\section{Declaration of competing interest}
The authors declare that they have no known competing financial interests or personal relationships that could have appeared to influence the work reported in this paper.

\section{CRediT Author Statement}
Yuqing Wang was involved in validation, visualization, and writing the original draft; Dong Xing contributed to validation, visualization, writing the original draft, as well as writing, reviewing, and editing the manuscript; Qika Lin, Wei Zhou, Hang Yang and Xianxun Zhu participated in validation and visualization.

\section{Acknowledgement}
This work was supported in part by Jilin Provincial Science and Technology Development Plan Project under Grant 20240302072GX.

\bibliographystyle{elsarticle-num}
\bibliography{xbib}

\end{document}